\documentclass[11pt,a4paper]{article}
\usepackage{times,latexsym}
\usepackage{url}
\usepackage[T1]{fontenc}

\usepackage[acceptedWithA]{tacl2018v2}

\usepackage{xspace,mfirstuc,tabulary}

\newif\iftaclinstructions
\taclinstructionsfalse % AUTHORS: do NOT set this to true
\iftaclinstructions
\renewcommand{\confidential}{}
\renewcommand{\anonsubtext}{(No author info supplied here, for consistency with
TACL-submission anonymization requirements)}
\newcommand{\instr}
\fi

\iftaclpubformat % this "if" is set by the choice of options

\else

\fi

\usepackage{booktabs} 
\usepackage{graphicx}
\usepackage{microtype}
\usepackage{amsmath}
\usepackage[utf8]{inputenc}
\usepackage{pdflscape}

\title{ParaNames: A Massively Multilingual Entity Name Corpus}

\author{Jonne S{\"a}lev{\"a} \and Constantine Lignos \\
  Michtom School of Computer Science\\
  Brandeis University\\
  \texttt{\{jonnesaleva,lignos\}@brandeis.edu} \\}

\begin{document}
\maketitle
\begin{abstract}

We introduce ParaNames, a multilingual parallel name resource consisting of 118 million names spanning across 400 languages.
Names are provided for 13.6 million entities which are mapped to standardized entity types (PER/LOC/ORG).
Using Wikidata as a source, we create the largest resource of this type to-date.
We describe our approach to filtering and standardizing the data to provide the best quality possible.
ParaNames is useful for multilingual language processing, both in defining tasks for name translation/transliteration and as supplementary data for tasks such as named entity recognition and linking.
We demonstrate an application of ParaNames by training a multilingual model for canonical name translation to and from English.
% and discuss the challenges and limitations of using a name resource specifically derived from Wikidata.
Our resource is released under a Creative Commons license (CC BY 4.0).\footnote{\url{https://github.com/bltlab/paranames}}

% Upon acceptance, we will release our code and entity name resource on GitHub.
\end{abstract}

\section{Introduction}

Our goal for ParaNames is to introduce a massively multilingual entity name resource that provides names for diverse types of entities in the largest possible set of languages and can be kept up to date through a nearly automated preprocessing procedure.
A large resource of names of this type can support development and improvement of multilingual language technology applications, as it is often important to know how real-world entities are represented across various languages.

The correspondences of names across languages are not always easy to model; they can involve a mix of transliteration and translation and often involve inconsistencies across languages or even among names in a given language.
As a concrete example, some country names are translated in Finnish, so \emph{United Kingdom} is written as \emph{Yhdistynyt kuningaskunta}, a literal, word-by-word, translation.
In contrast, smaller territories may or may not be translated: the U.S. states of \emph{North Carolina} and \emph{New York} are written as \emph{Pohjois-Carolina} (with \emph{North} translated) and \emph{New York}, respectively. 
Moreover, Finnish versions of the U.S. states are often idiosyncratically translated, e.g. California is represented as \emph{Kalifornia}, whereas Colorado is represented as \emph{Colorado}. 

The examples above demonstrate the complex choices that language speakers make in representing named entities---even when only dealing with Latin script---and underscore the need for a large-scale, multilingual resources of entity name correspondences to effectively model these phenomena.

Addressing this need is difficult.
Most research groups (ours included) lack the means to assemble annotators in hundreds of languages to produce a carefully manually curated resource with the coverage we desire.
But even if we had sufficient means, such a resource would quickly fall out of date and would be difficult to incrementally grow with time.

Our approach is to instead try to adapt an existing, continuously maintained data source to serve this purpose.
Our method of adapting the resource needs to be almost entirely automated to allow updates as the upstream data source is modified.
The data source itself needs to cover as broad a set of languages as possible, especially under-resourced ones.
And to have the most useful set of names in each language possible, we need to try to exercise proper quality control, for example ensuring that the entities in each language are in the desired script even when there are errors in the source data.

We selected Wikidata\footnote{\url{https://www.wikidata.org}} as our data source, as it is particularly suited for the task because of its wide coverage of entities and languages as well as its nature as a perpetually updating collection, one which enables continuous improvement and expansion.
In this paper, we present our approach to transforming the Wikidata knowledge graph into a dataset of person, location, and organization entities with parallel names.

However, our contribution lies not just in making this resource available.
We identify potential problems in the source data---such as the lack of standardization of the script(s) used in each language---and provide a processing pipeline that addresses them.
In addition to ensuring consistency in the scripts used for each language, we focus on making the names as parallel as possible by removing extraneous information that can accompany them.

The following sections describe the characteristics of our dataset and our approach to constructing it.
While our goal is to promote ParaNames as a useful resource, we examine the use of Wikidata from a skeptical perspective, pointing out properties that may limit its usefulness.

We plan to provide regular updates to this resource to include corrections and improvements to both Wikidata and our extraction process.
The Wikidata names we use as a source are CC0 (``no rights reserved'') licensed,\footnote{\url{https://www.wikidata.org/wiki/Wikidata:Copyright}} and our resource is licensed using the Creative Commons Attribution 4.0 International (CC BY 4.0) license.

% Following TACL submission rules, we cannot provide or link to the contents of the resource as supplementary material.
% However, we are committed to making all software and data related to this resource available via Github.

\section{Related work}
\label{sec:relatedwork}

% LATER: Cite "A Study of the Quality of Wikidata"

While there is previous work in the construction of multilingual name resources, we are not aware of an \emph{openly-accessible} resource containing the names of millions of \emph{modern} entities in many languages.

\citet{wu-etal-2018-creating} create a translation matrix of 1,129 biblical names, with each English name containing translations into up to 591 languages.
% They demonstrate that modern neural machine translation (NMT) can provide good performance in the task of transliterating these names into English.
% LATER: If dicussing modeling, add citation for moran-lignos-2020-effective

\citet{merhav-ash-2018-design} release bilingual name dictionaries for English and each of Russian, Hebrew, Arabic, and Japanese Katakana.
However, their resource is limited to a few languages and only covers single token person names.
In contrast, our dataset includes hundreds of languages, entities other than persons, and consists primarily of multi-token entity names.

The Named Entity Workshop (NEWS) shared task has created parallel name resources across a series of shared tasks.
In the 2018 version of the shared task \citep{chen-etal-2018-report,chen-etal-2018-news}, participants were asked to transliterate between language pairs involving English, Thai, Persian, Chinese, Vietnamese, Hindi, Tamil, Kannada, Bangla, Hebrew, Japanese (Katakana / Kanji), and Korean (Hangul), although the task did not include transliteration between all pairs.
The NEWS 2018 datasets are hand-crafted and much smaller than ours, at most 30k names per language pair.
Unlike our resource, the datasets for these shared tasks are not fully publicly available; the test set is held back and the each of the five training sets is subject to different licensing restrictions.

We do not claim to be the first to harvest the parallel entity names available from Wikidata or Wikipedia.
There is scattered prior work in this area, with one of the earliest explorations at scale being performed by \citet{irvine-etal-2010-transliterating}.
\citet{steinberger-etal-2011-jrc} also collected names for roughly 200,000 entities in 20 scripts and several languages, using Wikipedia and news articles as their data sources.
Building on their work, \citet{benites-etal-2020-translit} also used Wikipedia as a data source and automatically extracted potential transliteration pairs, combining their outputs with several previously published corpora into an aggregate corpus of 1.6 million names.
While all these works produced collections of entities that are more modern than those produced by e.g. \citet{wu-etal-2018-creating}, the total number of names is still far smaller than our present resource.

Specifically for lower-resourced languages, many approaches to named entity recognition and linking for the LORELEI program \citep{strassel-tracey-2016-lorelei} used Wikidata, Wikipedia, DBpedia, GeoNames, and other resources to provide name lists and other information relevant to the languages and regions for which systems were developed.
However, while ad-hoc extractions of these resources were integrated into systems, we are unable to identify prior attempts to create a transparent, replicable extraction pipeline and to distribute the extracted resources with wide language coverage.

\section{Data extraction and quality challenges}
\label{sec:data-extraction-and-quality-challenges}

To construct our dataset, we began by extracting all entity records from Wikidata and ingesting them into a MongoDB instance for fast processing.
Each entity in Wikidata is associated with several types of metadata, including a set of one or more names that different languages use to refer to it.
Given that we are working with such a large-scale dataset, there are important challenges that arise when working with the data, which we describe in this section.

\subsection{Language representation}

The number of languages that entities have labels in varies wildly across Wikidata.
For example, the entry for Alan Turing (\url{https://www.wikidata.org/wiki/Q7251}) will show his name written in over a hundred languages, including many that use non-Latin scripts.
Internally, each language is referred to using a language code.
However, many of the Wikimedia language codes that Wikidata uses do not correspond one-to-one with natural languages.\footnote{The relationship between Wikimedia language codes and other language codes is rather complex. Originally, the Wikimedia language codes were designed to comply with \href{https://datatracker.ietf.org/doc/html/rfc3066}{RFC3066}, but there are inconsistencies and \href{https://meta.wikimedia.org/wiki/Wiki_language_ISO_639-1_\%E2\%86\%92_BCP_47_proposal}{standardization is unlikely to occur soon}.
Some, but not all, of the language codes are identical to modern  \href{https://datatracker.ietf.org/doc/html/rfc5646}{BCP 47 codes (RFC5646)}.
In this paper, we try to distinguish between the Wikimedia language codes---which may identify a language along with a script, geographical region, or dialect---and higher-level language identifiers which use only first two letters of the language code. When we provide the total number of languages covered, we use the higher-level identifiers to prevent double-counting one language written using multiple scripts.}
Often there are several Wikimedia codes for a given spoken language, varying in script or geography.
For example, the Kazakh language is associated with the Wikimedia language codes \texttt{kk} (Kazakh), \texttt{kk-arab} (Kazakh in Arabic script), and \texttt{kk-latn} (Kazakh in Latin script).
These language codes can potentially be helpful in learning to transliterate between different scripts of the same language.
At other times, the language codes are specific to geography rather than writing system. In the case of Kazakh, there are three main geography-specific language codes: \texttt{kk-cn} (Kazakh in China), \texttt{kk-kz} (Kazakh in Kazakhstan) and \texttt{kk-tr} (Kazakh in Turkey).

In our analysis and the resource we distribute, if there is only a single name for a given language code across the entities we select, we do not include that name in our resource as having a single name would not constitute meaningful representation of the language.

\subsection{Script usage}
While language codes can identify a specific script for a language, unfortunately many Wikidata labels do not conform to the scripts used by each language.
In many cases, this is simply a data quality issue, such as with Greek where approximately 8.9\% of ORG entities are written in Latin script rather than the Greek alphabet.\footnote{We confirmed with a Greek speaker that this represented a data issue and not meaningful variation within the language about how names are written.}

However, in other cases, the presence of several scripts can also reflect real world-usage depending on the language, as many languages commonly use several scripts.
As an example, Kazakh uses both the Cyrillic and Arabic alphabets, thus multiple scripts are to be expected across a collection of names and our resource reflects this diversity.

\subsection{Providing entity types}

% removed point about per-type analysis of systems since we're not doing it
Even though entities often have detailed information about what they represent, Wikidata does not directly categorize entities as instances of higher-level types such as location (LOC), organization (ORG), and person (PER).
% ANSWER: does anyone even care about what we do at ingest time? 
% Even though downstream tasks often rely on this kind of type information, we opted not to perform any type inference for entities at ingest time. 
% Instead we chose to extract entity types dynamically when constructing the final name resource.
To obtain this information, we chose to extract entity types based on the Wikidata inheritance hierarchy when constructing our resource.
Specifically, we identified suitable high-level Wikidata types---Q5 (human) for PER, Q82794 (geographic region) for LOC, and Q43229 (organization) for ORG---and classified each Wikidata entity that is an instance of these types as the corresponding named entity type.

While the \texttt{instance-of} relation is transitive---i.e. all instances of a subtype are instances of the higher-level type---we noticed that taking all subtypes of these high-level types led to many entities that were not individual persons to be classified as PER, such as \textit{Government secretaries of Policies for Women of the State of Bahia} (\href{https://www.wikidata.org/wiki/Q98414232}{Q98414232}).
To exclude such entities, we required that PER entities must also explicitly be an instance of Q5 (person) in addition to any subclass types.

We did not observe similar problems for LOC and ORG entities, so we kept the typing rules unchanged for them.
If we had imposed a more stringent type requirement as we did for PER, it would decrease the number of entities by 3,075,536 for LOC (3,078,459 to 2,923) and 2,137,550 entities for ORG (2,196,303 to 58,753).
For PER the change in number of entities was relatively small (8,730,734 to 8,726,412).

As shown in Table~\ref{tab:entitytypes}, a relatively small number of entities get assigned to multiple types. While this is a result of multiple-inheritance in the entity type hierarchy of Wikidata, having multiple types is not incorrect as an entity can represent several different types. 
In our resource, we opted to preserve this information, as assigning only a single type to complex entities could make our dataset less useful by ignoring inherent entity typing uncertainty.

\begin{table}[tb]
\small
\centering
\begin{tabular}{lrr}
\toprule
\textbf{Entity type} &      \textbf{Count} & \textbf{Percentage} \\
\midrule
PER & 8,725,777 & 63.83\% \\
LOC & 2,747,869 & 20.10\% \\
ORG & 1,865,255 & 13.65\% \\
Mixed & 330,793 & $<$2.5\% \\
\midrule
\textbf{Total}           & 13,669,694 &    100.0\% \\
\bottomrule
\end{tabular}

    \caption{Number of entities and percentage of all entities assigned to each combination of LOC, ORG and PER in ParaNames.}
    \label{tab:entitytypes}
\end{table}

\begin{figure*}[tb]
\centering
\includegraphics[width=\linewidth]{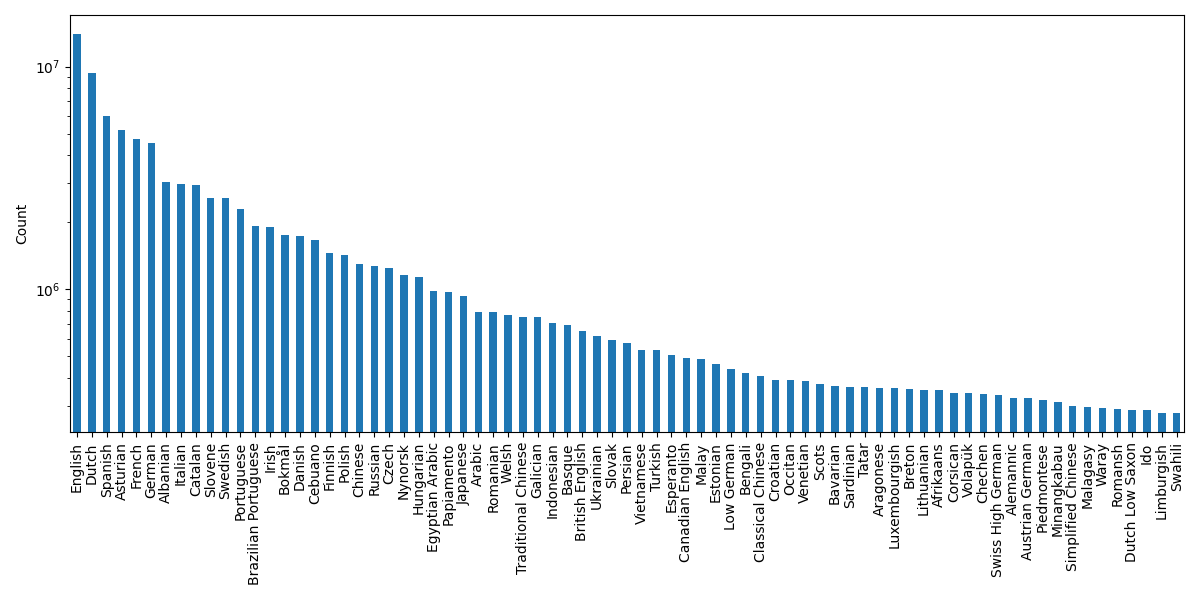}
\caption{Name counts across the 75 languages with the most names (languages identified by first two letters of Wikimedia language code, $\log_{10}$ y-axis).}
\label{fig:per_lang_barchart}
\end{figure*}

A visualization of the name counts for the 75 languages with the most entity labels in Wikidata is shown in Figure~\ref{fig:per_lang_barchart}.
The number of entity names in Wikidata varies greatly across languages, and counts are distributed according to a Zipf-like power law distribution where a few languages contain most of the names.
As expected, many of the largest languages are also large in terms of number of speakers. However, there are notable exceptions, such as Asturian, which contains the fourth largest number of entity names, despite only having fewer than a million native speakers. We suspect that this is an artifact of non-human editing on Wikipedia, and many of these entities appear to be copies of the English name.
We discuss this further in Section~\ref{sec:limitations}.

The relative proportions of entity types also seem to vary, with PER entities comprising the bulk of names for most languages. 
There are exceptions, however. For instance, in Ukrainian, LOC entities account for approximately 45\% of the names, which is substantially larger than the approximately 22\% of English entities which are of type LOC.

\section{Improving data quality}

To ensure our resource is of the highest quality possible, we identified two properties that all languages in our corpus should adhere to for maximal usefulness.
First, for each language, all entities in a language should be written in script(s) that match its real-world usage.
Second, parallel names in our corpus would ideally have the same information on both sides; additional information like titles that appear in one language and not the other should be removed.

\subsection{Script standardization}

For the first property, we chose to normalize the names for each language by filtering out names that are not in the desired script(s) for the language.
An example of this would be a Russian entity label like \emph{Canada} which is not written in Cyrillic.

While we explored automated methods of doing this, ultimately we decided that manually constructing a list of allowed scripts for each language would yield the best results.
For each language, we used Wikipedia as an authoritative source to look up which scripts are used to write the language, and filtered out all names whose most common Unicode script property is not among the allowed ones.
We used the PyICU library\footnote{\url{https://gitlab.pyicu.org/main/pyicu}} to identify the most frequent Unicode script tag in each name based on individual characters.

To quantify how much this filtering changed the entity names associated with each language, we attempted to measure script uniformity for each language.
For each language, we aggregated the Unicode script tags produced by PyICU across names for each language and computed the entropy of this distribution, calling this quantity \emph{script entropy} and used it as a proxy for script consistency within a language's names.
Languages whose names are consistently written in a single script will have near-zero entropy.

The filtering process decreased the average script entropy from 0.142 to 0.022.
After filtering, 463 Wikimedia language codes remained with a total of 118,894,875 names across 13,669,694 entities.

\subsection{Matching information across languages}

We observed that some names contain additional information in parentheses following the actual tokens of the entity name, intended to help disambiguate the name from other similar-looking entities.
For instance, the entity with the English label \emph{Wang Lina (boxer)} (\href{https://www.wikidata.org/wiki/Q60834172}{Q60834172}) has a Russian label which contains the translation of word \emph{boxer} in parentheses.
However, this is not the case for all languages: for example, the Spanish name for the entity is simply \emph{Wang Lina}.

To standardize the amount of information per name across languages, we remove all parentheses and tokens inside them using a regular expression.

\section{Limitations}
\label{sec:limitations}

\subsection{Single name per language code}
\label{subsec:singlename}
Our dataset only uses the ``label'' property in Wikidata to identify names for entities.
One of the potential limitations of this approach is that a given entity can only have a single label within a single Wikimedia language code, even though there may be multiple possible transliterations of an entity name for that language code.
This can be especially problematic for languages that use more than one script but for which a finer-grained language code the specifies the script, such as \texttt{sr-cyrl}, is not available.
For example, Bosnian only has the language code \texttt{bs} but is commonly written in Cyrillic and Latin scripts.
% Since our resource is meant to undergo regular updates, we chose to not tackle this issue directly for now.
% Instead, alternate names in orthographies that are valid for a language but previously absent from Wikidata can be included in future releases of our dataset.

There is a possible solution in Wikidata for this limitation.
There is an ``also-known-as'' (AKA) property, which for many entities contains useful examples of real-world names used to refer to it and can include alternative transliterations.
Unfortunately, it often includes names that only loosely correspond to the canonical name of the entity.
For example, AKAs for the late U.S. Supreme Court justice Ruth Bader Ginsburg (\href{https://www.wikidata.org/wiki/Q11116}{Q11116}) contain not only her full name, \emph{Ruth Joan Bader Ginsburg}, but also common aliases from popular culture, such as \emph{Notorious RBG}.  
In the case of Donald Trump (\href{https://www.wikidata.org/wiki/Q22686}{Q22686}) the AKAs contain other variations of his name (\emph{Donald John Trump}, \emph{Donald J. Trump}, etc.), but also pseudonyms that he has used that do not correspond to his actual name (\emph{John Barron}, \emph{John Miller}, \emph{David Dennison}, etc.).
While this information could be argued to be useful for downstream tasks such as entity linking, we felt that these alternative names introduced potentially unwanted variation in the names across languages.
For this reason, we chose not to include the also-known-as fields in our dataset at this time.

There are other datasets that do not share the limitation of only having one name for an entity per language.
For example, the NEWS 2018 shared task dataset \citep{chen-etal-2018-report,chen-etal-2018-news} allows for multiple correct reference transliterations.
Participants in that shared task also produced a ranked list of candidate translations, which can help handle the arbitrary nature of picking from an otherwise synonymous list of candidates.

\subsection{Wikidata quality issues}
Another limitation of our resource is our limited ability to address cases where Wikidata contains labels that may have been copied from one language to another without scrutiny.
While our preprocessing pipeline removes names that appear in an incorrect script for a given language---for example, a Latin-script named copied into a language that does not use the Latin script---names blindly copied from one language into another that are in the correct script cannot reliably be detected.

Thus, a Latin-script language like Asturian which contains many names on Wikidata but has few speakers---raising the question of whether those names were added by actual speakers of the language---may have many names in our resource that were copied from English without any human review.
We cannot automatically filter out these names, and collecting native speaker judgments on each one would be cost-prohibitive.
While heuristic approaches like computing the percentage of names exactly equal to English could be employed, as many names are identical across languages, this may not be a meaningful heuristic.

\subsection{Nicknames}
Another source of variation not addressed in this work is nicknames, which can create non-parallelism.
For example, while the English Wikidata label for \emph{Joe Biden} uses the nickname \emph{Joe}, a minority of the labels in other languages use forms of \emph{Joseph}.

However, it can be difficult to differentiate the use of nicknames from ordinary transliteration of the full name, which may show the affects of phonological adaptation or morphological simplification.
For example, the first name of \emph{Konstantinos Ypsilantis} (\href{https://www.wikidata.org/wiki/Q2272090}{Q2272090}) may be written with the nominative \emph{-os} suffix of the original Greek in some languages but appear without it in others (Polish: \emph{Konstantyn}, Slovenian: \emph{Konstantin}, etc.).

Unlike removing undesirable nickname variation like \emph{Joe/Joseph} and \emph{Will/William}, normalizing the dataset to always include or remove the \emph{-os} suffix in the name of cross-language consistency would overly simplify the translation task.

\section{Experimental setup}
\label{sec:experimental-setup}

\begin{table}[tb]
\small
    \centering
    \begin{tabular}{llrr}
    \toprule
      Language &                    Script &  Names & \% Train \\
    \midrule
        Arabic &                    Arabic & 500,000 &      11.1\% \\
      Japanese &         Kanji, Kana$^{*}$ & 500,000 &      11.1\% \\
      Swedish &                      Latin & 500,000 &      11.1\% \\
      Russian &                   Cyrillic & 500,000 &      11.1\% \\
      Persian &                     Arabic & 457,200 &      10.2\% \\
    Vietnamese &                     Latin & 429,185 &       9.6\% \\
    Lithuanian &                     Latin & 282,074 &       6.3\% \\
        Hebrew &                    Hebrew & 205,704 &       4.6\% \\
        Korean &                    Hangul & 203,042 &       4.5\% \\
      Latvian &                      Latin & 177,577 &       4.0\% \\
      Armenian &                  Armenian & 161,957 &       3.6\% \\
         Greek &                     Greek & 149,515 &       3.3\% \\
        Kazakh &                  Cyrillic & 124,574 &       2.8\% \\
          Urdu &                    Arabic & 103,803 &       2.3\% \\
          Thai &                      Thai &  72,112 &       1.6\% \\
      Georgian &                  Georgian &  70,965 &       1.6\% \\
         Tajik &          Cyrillic,  Latin &  52,574 &       1.2\% \\
    \midrule
    Total &                                &         &     100.0\% \\
    \bottomrule
    \end{tabular}
    \caption{Parallel training data statistics and the script(s) used to write the names in our dataset. The development and test sets were each balanced to 5,000 names per language. $^{*}$Kana jointly refers to the two Japanese syllabaries Hiragana and Katakana.}
    \label{tab:parallel_data_stats}
\end{table}

\subsection{Task definition}

To demonstrate an application of ParaNames, we use it to train models that translate entity names from many languages to English and from English to many languages.
We call this task \emph{canonical name translation}, as the task is to translate the Wikidata label (canonical name) for an entity into the label in another language.

It is important to clarify what this task is and what it is not.
We do not refer to this task as name transliteration because not every name pair is strictly transliterated; often the mapping includes elements of transliteration, translation (especially for organization names), and sometimes morphological inflection/deinflection as well.
The task is also not the translation of a name within a sentence, which often requires correct morphological inflection of the name in its sentential context.

\subsection{Data selection}

For our experiments, we translate named entities from 17 languages---Arabic, Armenian, Georgian, Greek, Hebrew, Japanese, Kazakh, Korean, Latvian, Lithuanian, Persian (Farsi), Russian, Swedish, Tajik, Thai, Vietnamese, and Urdu---into English and vice versa, using a single multilingual model for each translation direction.

We chose these languages as they cover a wide geographic distribution,\footnote{Unfortunately, we were not able to achieve quite as wide geographic distribution as we hoped because we were unable to find an African language with useful data for this task in our resource. All the Latin-script African languages that we explored had almost all of their names identical to English, and the non-Latin script languages had too few names.} as well as several different orthographic systems, language families and typological features.
While there is overlap between the languages in terms of scripts, some, such as Tajik and Persian, are often considered closely related despite using different scripts.

While many entity labels for Latin-script languages are identical to the English label, this is not always the case. For example, Vietnamese relies heavily on diacritics, and English names are often spelled phonetically and inflected when written in Latvian (e.g. \textit{Dzo Baidens} for Joe Biden).
By including a small number of Latin-script languages in our experiments, we are able to assess our model's performance on such languages without overly inflating performance numbers by having a large part of the evaluation set consist of names written identically to English.

The languages we selected also have varying amounts of data available in our resource.
All the languages we selected had sufficient names to allow for the development and test sets to be equally balanced across languages (5k names per language), but there was an order of magnitude difference between the language with the fewest names available for the training data (Tajik, ~50k) and those which we limited to 500k names in training (Arabic, Japanese, Russian, Swedish) to avoid oversampling.
Due to limits in the computational resources available to us, we are not able to perform experiments in a larger number of languages; however, we believe this to be a representative set for the purpose of demonstrating the tasks that can be defined using ParaNames.

\paragraph{Data splitting}
To create the parallel data for this task, we extracted all Wikidata IDs that had names in English and at least one of the other languages in our selected set.
We divided the Wikidata IDs to either the train, development, or test set using an 80/10/10 split.
The overall statistics of the parallel data can be seen in Table~\ref{tab:parallel_data_stats}.
While per-ID splitting does not guarantee identical language stratification across train, development, and test sets, we employ it to avoid a data leakage scenario where the English side of a given entity name might appear in more than one of our train, development, or test sets.
Notably, this leakage does occur in the data split created by \citet{wu-etal-2018-creating} because they split the data by source-target name pairs, not whole entities.
To further balance our datasets and avoid overly biasing our models towards the higher-resourced languages, we also capped the maximum number of names in our splits. For training data, we allowed up to 500,000 pairs, whereas for development and test, we set a limit of 5,000 names.

\paragraph{Lack of manual annotation}
The test set used in evaluation was extracted directly from our resource and no additional cleaning or manual annotation was performed, except for script standardization and parenthesis removal, as outlined in Section~\ref{sec:data-extraction-and-quality-challenges}.
We believe this approach to be reasonable, as script standardization only filters out names but does not alter those which are included. 
Most of the languages featured in our experiments use a non-Latin script and the prevalence of entities in incorrect scripts was the largest data quality issue.
By minimizing the amount of manual intervention, we also maximize the extent to which our experiments correspond to translating between names that have been produced through actual usage of the language, as opposed to heuristics.

\paragraph{Special tokens} After creating the data splits, we augmented the source side of each name pair with a ``special token'' that indicates information about the non-English language.
In the case of X $\rightarrow$ En models, this corresponds to the source name and in En $\rightarrow$ X models the target name.
The purpose of the special token is to help our model better manage the multilingual training setting by keeping languages separate, especially ones with potentially overlapping scripts such as Tajik, Russian and Kazakh or Swedish, Latvian, and Lithuanian.
We experimented with what information to include in the special token(s); details are given in Section~\ref{sec:results}.

\subsection{Evaluation metrics} 
We evaluate using three metrics: 1-best accuracy (where each a name translation must match the reference \emph{exactly}), \emph{character error rate} (CER), computed analogously to word error rate but at the character level, and \emph{mean F1-score} based on longest common subsequence \citep{chen-etal-2018-news}. 

Our use of mean F1-score is motivated by its use in the original NEWS 2018 Shared Task on Machine Transliteration, the most similar shared task to our experiments. 
The authors define the mean F1-score as the average of the individual F1-scores of each candidate-reference pair.
The F1-score of individual candidate-reference pairs is defined the usual way, with precision and recall computed using the longest common subsequence:

%\small
\begin{align}
    \text{LCS}(C, R) &= |C| + |R| - \text{ED}(C, R) \\
    \text{Precision}(C, R) &= \frac{\text{LCS}(C, R)}{|C|} \\
    \text{Recall}(C, R) &= \frac{\text{LCS}(C, R)}{|R|}
\end{align}
%\normalsize

Intuitively, LCS measures the overlap between candidate and reference strings, computed in a way that accounts for character order.
This ensures that pairs that are anagrams of each other do not receive high scores even though they overlap completely in terms of unordered characters.

% DONE: This isn't very clear, and the equation that's called out is just F1, when we should be focused on how the precision and recall are defined (which comes later)
% On the other hand, the F1-score can also be understood as the intersection of two fuzzy sets. 
% Concretely, let $C$ and $R$ be candidate and reference words, and define the fuzzy sets $S_{R|C}$ and $S_{C|R}$ using the membership functions $\mu_{R|C} = \text{prec}(C,R)$ and $\mu_{C|R} = \text{rec}(C, R)$, where $\text{prec}$ and $\text{rec}$ refer to precision and recall, respectively.
% Noting that $F_1: [0, 1] \times [0, 1] \rightarrow [0,1]$ satisfies all the axioms of a fuzzy intersection\footnote{For more details we refer the reader to \url{https://en.wikipedia.org/wiki/Fuzzy_set_operations}.}, we define $S_{R|C} \cap S_{C|R}$ using the membership function

% \begin{align}
%     \mu_{S_{R|C} \cap S_{C|R}} &= F_1(C, R) \\
%                              &=\frac{2\cdot \text{prec}(C,R)\cdot \text{rec}(C,R)}{\text{prec}(C,R) + \text{rec}(C,R)}
% \end{align}

% For a given pair $(C, R)$, the membership function $\mu_{S_{R|C} \cap S_{C|R}}$ quantifies how likely the pair is to be in the set $S_{R|C} \cap S_{C|R}$, and a high value indicates the pair is likely to have both high precision and recall.
% As the mean F1-score is simply the regular F1-score averaged over all $(C, R)$ pairs in the corpus, it can be understood as estimating the degree of fuzziness for an arbitrary pair.

\subsection{Model details}

The model we use is a simple character-level Transformer-based translation model \textit{trained from scratch}.
We use the model structure and hyperparameters from past transliteration experiments by \citet{moran-lignos-2020-effective} with minor changes.
We use a 4-layer Transformer with a hidden layer size of 1024, embedding dimension of 200, 8 attention heads, and a learning rate of 0.0003, with a dropout probability of 0.2. The label smoothing parameter is set to 0.1, and batch size is set to 128.
We use the Adam optimizer for a maximum of 75,000 updates. Each experiment is repeated 5 times using random seeds ranging from 1917 to 1921. 
A single NVIDIA RTX 3090 GPU is used for both training and decoding.
For each experimental condition (i.e. direction, source-side special token setting, and language), training the model took roughly 9 hours and evaluation took roughly 15-30 minutes.
We implement our model using fairseq \citep{ott-etal-2019-fairseq}.

\section{Results}
\label{sec:results}
% TODO: analysis of what lang-specific properties make it hard

We first performed a baseline experiment using a source-side special token that only conveys the language being translated into (for English to all languages) or out of (for all languages to English).
We then performed a second set of experiments where we modified the information contained in special tokens to assess the effects on performance.

Results reported in all tables are the mean value and the standard deviation of the mean (standard error) computed across training five models with different random seeds.
All values have been rounded.
Accuracy and F1 are reported out of 100 points for readability.
For CER, 1.0 reflects a 100\% error rate (lower scores are better).

\subsection{Language-only special token baseline}

\begin{table*}[tb]
\small
\centering
\begin{tabular}{llrrrrr}
\toprule
   & \multicolumn{3}{c}{X $\rightarrow$ En} & \multicolumn{3}{c}{En $\rightarrow$ X} \\
   \cmidrule(lr){2-4} \cmidrule(lr){5-7}
  Language & Accuracy & CER & F1 & Accuracy & CER & F1 \\
\midrule
   Swedish &    88.25 $\pm$ .02 &  0.08 $\pm$ .00 & 97.15 $\pm$ .01 &    85.60 $\pm$ .04 &  0.10 $\pm$ .00 & 96.11 $\pm$ .02 \\
Vietnamese &    80.75 $\pm$ .02 &  0.17 $\pm$ .00 & 94.08 $\pm$ .01 &    48.86 $\pm$ .01 &  0.35 $\pm$ .00 & 82.87 $\pm$ .01 \\
   Latvian &    67.86 $\pm$ .02 &  0.14 $\pm$ .00 & 95.19 $\pm$ .01 &    69.28 $\pm$ .07 &  0.13 $\pm$ .00 & 95.49 $\pm$ .01 \\
    Kazakh &    55.38 $\pm$ .04 &  0.16 $\pm$ .00 & 93.93 $\pm$ .01 &    58.69 $\pm$ .09 &  0.14 $\pm$ .00 & 94.85 $\pm$ .02 \\
     Tajik &    49.62 $\pm$ .05 &  0.20 $\pm$ .00 & 92.77 $\pm$ .01 &    54.38 $\pm$ .02 &  0.18 $\pm$ .00 & 93.82 $\pm$ .02 \\
Lithuanian &    47.39 $\pm$ .03 &  0.28 $\pm$ .00 & 89.53 $\pm$ .01 &    50.76 $\pm$ .09 &  0.23 $\pm$ .00 & 91.61 $\pm$ .03 \\
      Thai &    43.94 $\pm$ .05 &  0.29 $\pm$ .00 & 89.91 $\pm$ .01 &    14.80 $\pm$ .04 &  0.42 $\pm$ .00 & 83.01 $\pm$ .02 \\
  Armenian &    39.92 $\pm$ .05 &  0.28 $\pm$ .00 & 90.04 $\pm$ .01 &    50.45 $\pm$ .05 &  0.22 $\pm$ .00 & 92.41 $\pm$ .01 \\
  Georgian &    34.44 $\pm$ .02 &  0.29 $\pm$ .00 & 89.29 $\pm$ .01 &    51.82 $\pm$ .04 &  0.22 $\pm$ .00 & 92.56 $\pm$ .01 \\
    Korean &    33.27 $\pm$ .05 &  0.32 $\pm$ .00 & 88.46 $\pm$ .01 &    38.63 $\pm$ .05 &  0.33 $\pm$ .00 & 88.18 $\pm$ .01 \\
   Russian &    32.81 $\pm$ .06 &  0.38 $\pm$ .00 & 84.80 $\pm$ .02 &    44.59 $\pm$ .04 &  0.33 $\pm$ .00 & 89.81 $\pm$ .02 \\
      Urdu &    31.92 $\pm$ .03 &  0.23 $\pm$ .00 & 91.48 $\pm$ .01 &    14.14 $\pm$ .08 &  0.45 $\pm$ .00 & 80.74 $\pm$ .03 \\
  Japanese &    29.00 $\pm$ .04 &  0.33 $\pm$ .00 & 87.79 $\pm$ .01 &    28.70 $\pm$ .01 &  0.42 $\pm$ .00 & 84.42 $\pm$ .02 \\
   Persian &    28.68 $\pm$ .05 &  0.28 $\pm$ .00 & 89.84 $\pm$ .02 &    22.90 $\pm$ .05 &  0.41 $\pm$ .00 & 81.64 $\pm$ .05 \\
    Arabic &    25.74 $\pm$ .03 &  0.32 $\pm$ .00 & 89.23 $\pm$ .01 &    41.70 $\pm$ .02 &  0.28 $\pm$ .00 & 89.40 $\pm$ .01 \\
     Greek &    24.70 $\pm$ .03 &  0.35 $\pm$ .00 & 86.60 $\pm$ .01 &    29.67 $\pm$ .06 &  0.36 $\pm$ .00 & 86.88 $\pm$ .01 \\
    Hebrew &    15.24 $\pm$ .07 &  0.44 $\pm$ .00 & 84.58 $\pm$ .02 &    35.71 $\pm$ .03 &  0.34 $\pm$ .00 & 88.16 $\pm$ .01 \\
\midrule
   Overall &    42.88 $\pm$ .02 &  0.27 $\pm$ .00 & 90.27 $\pm$ .01 &    43.57 $\pm$ .02 &  0.29 $\pm$ .00 & 88.94 $\pm$ .01 \\
\bottomrule
\end{tabular}
\caption{Canonical name translation performance on the test set using our baseline configuration with language special tokens on the source side, sorted by descending accuracy for the X $\rightarrow$ En task.}
\label{baseline-results}
\end{table*}

As our first experiment, we evaluated canonical name translation performance in both En $\rightarrow$ X and X $\rightarrow$ En directions using language special tokens on the source side.
The overall results for both translation directions, computed on the test set, are given in Table \ref{baseline-results}. 
The last row (``Overall'') gives micro-averaged performance across all languages.
% MAYBE: macro-average

\paragraph{X $\rightarrow$ English}
When translating to English, our model performs best on Swedish and Vietnamese, with 1-best accuracy in the 80-90\% range for both languages. 
This is unsurprising, as both languages use the Latin script and contain many names spelled identically to English.
Immediately following them is Latvian, where accuracy is lower as many names need to be inflected and the names generally match English less often.

Kazakh and Tajik, both written in the Cyrillic script, immediately follow Swedish and Vietnamese, which makes sense as well since Cyrillic can be transliterated to Latin script relatively unambiguously.
Russian, on the other hand, seems to perform considerably worse than the other Cyrillic-script languages, perhaps due to names being longer in Russian and the use of patronymics.

Model performance is consistently worst on Hebrew.
The most likely cause is lack of vowels in the Hebrew names, which the model must infer when translating to English.

When qualitatively inspecting model outputs, we noticed that often our model relies too heavily on transliteration when some words must be translated or vice versa.
Many outputs were also incorrect because they lacked extra information that was only present on the target side and omitted on the source side. 
For example, tokens like \textit{Stream} in \textit{Cuiva Stream} (\href{https://www.wikidata.org/wiki/Q21412684}{Q21412684}) are only present in the English name and cannot be learned by seeing the non-English source label.

\paragraph{English $\rightarrow$ X}
% As Figure~\ref{fig:en2all-swarmplot} shows, the English $\rightarrow$ X model performs better than the X $\rightarrow$ English based on median accuracy and character error rate. 
When translating from English, the performance rankings of the top languages are similar to when translating to English.
Swedish and Latvian have the highest accuracy, followed by Kazakh, Tajik, and Georgian.
We again find that the model performs worse on Russian than other languages that use the Cyrillic script.

For Hebrew, the model performs much better than when translating from English, as it does not have to infer the vowels, only delete them.
For Thai, the reverse is true and the English-Thai direction performs significantly worse than Thai-English. 
Since the Thai script indicates vowels using combining diacritics, we hypothesize this might be more difficult for the model to get exactly correct than English where vowels are written out explicitly.
This might be improved by experimenting with different forms of Unicode normalization, which we did not utilize in our experiments.

\paragraph{Metrics}
While CER and accuracy show broad separation across the languages, mean F1-score is always above 80, even in cases when the accuracy is low and CER is high.
For example, Hebrew to English translation has an accuracy of 15\%, a CER of .44, but an F1-score of 84.58.
While we report the mean F1-score metric here for completeness because it was used in the best-known transliteration shared task, our results suggest that it may be the least discriminating of the metrics we use.

\subsection{Finding the optimal special tokens}
\label{sec:tag-ablation-experiments}

\begin{table*}[tb]
\small
\centering
\begin{tabular}{llrrrrr}
\toprule
   & \multicolumn{3}{c}{X $\rightarrow$ En} & \multicolumn{3}{c}{En $\rightarrow$ X} \\
   \cmidrule(lr){2-4} \cmidrule(lr){5-7}
  Language & Accuracy & CER & F1 & Accuracy & CER & F1 \\
\midrule
   Swedish &          88.17 $\pm$ .03 &           0.08 $\pm$ .00 &          97.13 $\pm$ .01 &          85.69 $\pm$ .02 &           0.10 $\pm$ .00 &          96.13 $\pm$ .01 \\
Vietnamese &          80.96 $\pm$ .01 &           0.17 $\pm$ .00 &          94.11 $\pm$ .01 &          48.78 $\pm$ .03 &           0.35 $\pm$ .00 &          82.89 $\pm$ .01 \\
   Latvian & \textbf{68.64 $\pm$ .03} &           0.14 $\pm$ .00 & \textbf{95.28 $\pm$ .01} & \textbf{70.50 $\pm$ .04} & \textbf{ 0.12 $\pm$ .00} & \textbf{95.79 $\pm$ .00} \\
    Kazakh & \textbf{56.33 $\pm$ .05} &           0.16 $\pm$ .00 & \textbf{94.01 $\pm$ .00} &          59.78 $\pm$ .07 &           0.13 $\pm$ .00 &          95.00 $\pm$ .01 \\
     Tajik & \textbf{50.36 $\pm$ .03} &           0.20 $\pm$ .00 &          92.84 $\pm$ .01 & \textbf{54.83 $\pm$ .04} & \textbf{ 0.17 $\pm$ .00} & \textbf{93.99 $\pm$ .01} \\
Lithuanian & \textbf{48.06 $\pm$ .06} & \textbf{ 0.27 $\pm$ .00} & \textbf{89.72 $\pm$ .01} & \textbf{54.21 $\pm$ .05} & \textbf{ 0.20 $\pm$ .00} & \textbf{92.62 $\pm$ .02} \\
      Thai & \textbf{45.38 $\pm$ .07} & \textbf{ 0.28 $\pm$ .00} & \textbf{90.12 $\pm$ .02} &          15.07 $\pm$ .05 &           0.41 $\pm$ .00 & \textbf{83.43 $\pm$ .02} \\
  Armenian &          40.70 $\pm$ .08 & \textbf{ 0.27 $\pm$ .00} & \textbf{90.14 $\pm$ .01} & \textbf{51.78 $\pm$ .06} &           0.21 $\pm$ .00 &          92.43 $\pm$ .01 \\
  Georgian & \textbf{35.56 $\pm$ .05} & \textbf{ 0.29 $\pm$ .00} &          89.40 $\pm$ .01 & \textbf{53.13 $\pm$ .05} &           0.22 $\pm$ .00 & \textbf{92.68 $\pm$ .01} \\
    Korean & \textbf{35.20 $\pm$ .04} & \textbf{ 0.31 $\pm$ .00} & \textbf{88.98 $\pm$ .01} & \textbf{39.29 $\pm$ .04} &           0.33 $\pm$ .00 & \textbf{88.34 $\pm$ .01} \\
   Russian &          32.92 $\pm$ .04 &           0.38 $\pm$ .00 &          84.82 $\pm$ .03 & \textbf{45.68 $\pm$ .03} & \textbf{ 0.32 $\pm$ .00} &          89.94 $\pm$ .01 \\
      Urdu & \textbf{32.74 $\pm$ .04} & \textbf{ 0.22 $\pm$ .00} & \textbf{91.62 $\pm$ .01} &          14.23 $\pm$ .06 &           0.45 $\pm$ .00 &          80.76 $\pm$ .03 \\
  Japanese &          29.53 $\pm$ .04 & \textbf{ 0.32 $\pm$ .00} &          87.90 $\pm$ .01 &          28.61 $\pm$ .04 & \textbf{ 0.42 $\pm$ .00} & \textbf{84.64 $\pm$ .01} \\
   Persian & \textbf{29.47 $\pm$ .04} &           0.27 $\pm$ .00 &          89.92 $\pm$ .02 &          22.60 $\pm$ .07 &           0.42 $\pm$ .00 &          81.81 $\pm$ .05 \\
    Arabic & \textbf{27.07 $\pm$ .05} & \textbf{ 0.31 $\pm$ .00} & \textbf{89.51 $\pm$ .01} &          41.67 $\pm$ .03 &           0.28 $\pm$ .00 &          89.33 $\pm$ .01 \\
     Greek & \textbf{25.74 $\pm$ .08} & \textbf{ 0.35 $\pm$ .00} & \textbf{86.81 $\pm$ .01} &          30.18 $\pm$ .03 &           0.36 $\pm$ .00 &          86.88 $\pm$ .01 \\
    Hebrew & \textbf{16.34 $\pm$ .03} & \textbf{ 0.42 $\pm$ .00} & \textbf{84.89 $\pm$ .01} &          36.00 $\pm$ .03 & \textbf{ 0.33 $\pm$ .00} & \textbf{88.23 $\pm$ .01} \\
\midrule
   Overall & \textbf{43.72 $\pm$ .02} & \textbf{ 0.27 $\pm$ .00} & \textbf{90.42 $\pm$ .00} & \textbf{44.24 $\pm$ .01} &           0.29 $\pm$ .00 & \textbf{89.11 $\pm$ .00} \\
\bottomrule
\end{tabular}
\caption{Canonical name translation performance of the best special token configuration.
For X $\rightarrow$ En, the best configuration is having language, type, and script information, and for En $\rightarrow$ X, it is having language and type information.
Languages are sorted by descending accuracy on the X $\rightarrow$ En side.
Boldface indicates statistically significant performance differences from the language-only special token baseline.}
\label{best-results}
\end{table*}

\begin{figure*}[tb]
    \centering
    \includegraphics[width=\linewidth]{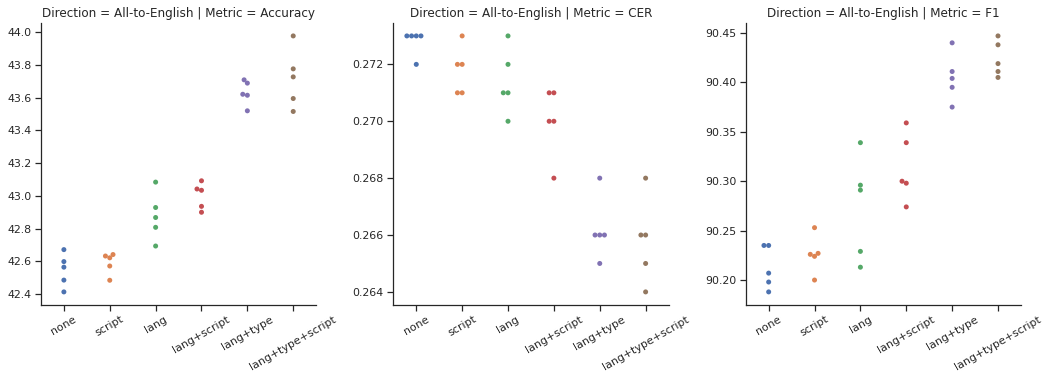}
    \caption{Canonical name translation performance across special token conditions for the X $\rightarrow$ English direction.}
    \label{fig:all2en-swarmplot}
\end{figure*}

\begin{figure*}[tb]
    \centering
    \includegraphics[width=\linewidth]{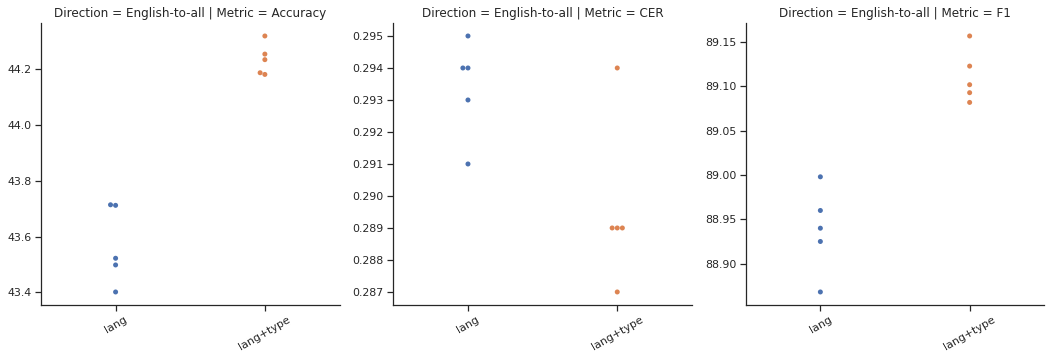}
    \caption{Canonical name translation performance across special token conditions for the for English $\rightarrow$ X direction.}
    \label{fig:en2all-swarmplot}
\end{figure*}

In addition to adding source-side language tokens to our parallel data, we also hypothesized that incorporating other kinds of information could be helpful.
Entity type information can potentially be helpful in guiding the model decoder, as the canonical name translation task may vary depending on the type of entity being translated.
In general, most person names are transliterated while organization names tend to include more translation, and many location name pairs contain tokens on one side that are absent from the other.
Script information can also be useful when dealing languages that are written in several scripts or to help encourage transfer across languages that share a script.

To investigate these hypotheses, we repeated Experiment 1 using various different kinds of special token settings: a language token (\texttt{<ru>}) in conjunction with either a type token (\texttt{<PER>}), a script token (\texttt{<Cyrillic>}), or both.
We also performed an ablation experiment by removing special tokens when possible.

Entity type tokens were generated from the PER/LOC/ORG type information in our resource inferred from Wikidata types.
For the small number of entities that mapped to multiple types, an arbitrary one was chosen.
Script tokens were generated using the PyICU library as with script filtering.
For each name, the special token reflected the most frequent Unicode script used in that particular name (not necessarily in the language in general).

For the X $\rightarrow$ English direction, we experimented with the following special token configurations: no special token; script only; language only; language and script; language and entity type; language, entity type, and script. 
For English $\rightarrow$ X we only evaluated having a language token and language and entity type tokens, as fewer configurations were possible.
The language token must always be present for the model to know what language to translate into, so we did not experiment with removing it.
We could not use the script token for English $\rightarrow$ X as it is computed from the non-English (target) side of the translation; using it would effectively leak specific information about the test data as part of the model's job is to predict which script to use in the case of a language that uses multiple scripts.
This leakage is relevant since each entity in our dataset only has a single name per language (see Section~\ref{subsec:singlename}). % THINK: Good enough?

The full results of our experiments across all languages and special token settings can be seen in Table~\ref{bigtable-all-results}.
A more interpretable visualization of the data is given in Figures \ref{fig:all2en-swarmplot} and \ref{fig:en2all-swarmplot} which contain ``swarm plot'' visualizations of our results across special token conditions and different metrics.
These enable viewing of all data points for overall performance across languages; the spread of points for each configuration gives the variation due to the different random seeds used in training.

For the X $\rightarrow$ English direction, we can see that using no special token or a script-only special token perform similarly, and there are clear improvements from adding language and entity type special tokens.
There appears to be some marginal improvement from adding the script special token to language-only and language and entity type settings.
For the English $\rightarrow$ X direction, we can see that adding entity type information on top of the language provides a clear improvement.

Regardless of metric, the differences are relatively small in our ablation study.
When translating to English, the overall accuracy when no special token is used is 42.55, adding a language-only special token increases that to 42.88, and adding type information on top of that increases it to 43.63 (Table~\ref{bigtable-all-results}).
Most languages behave similar to the overall trend, even though the effects vary slightly across languages. 
At one extreme, the average improvement for Korean when using language and type tokens compared to the language tag -only baseline is 1.93, substantially larger than the micro-averaged improvement of 0.75. 
On the other hand, for Swedish the mean change from baseline is -0.08, which is significantly lower than the micro-averaged change.
Overall we would have predicted that the language special token would have more impact than entity type information.
This underscores the importance of using an entity type special token for this task.
The limited usefulness of the script token suggests that our model is already able to determine the necessary script information via the language special token, and that the benefit from additional script information is marginal.
This can be explained by the fact that most languages we work with consist of names written in only a single script. As a result, given a language, the script is trivial for the model to deduce.

% Figure~\ref{fig:all-metrics-boxplot} displays a box-and-whiskers plot visualization of all of our results. Each box represents the range between the 25th and 75th percentiles, also known as the \textit{interquartile range} (IQR). The line inside the box indicates the median and the whiskers of the plot extend to 1.5 times the IQR. Any individual points outside the whiskers are considered outliers. 

Table~\ref{best-results} shows the results on our best special token setting.
As the results within a language tend to be quite similar across special tokens, we performed statistical significance testing to assess whether there are differences between the various special token settings.     
For each language, metric, and translation direction, we performed a two-tailed Mann-Whitney U test, which is a nonparametric alternative to the two-sample $t$-test and requires no assumptions about the distribution of the data. 
For each test, we compared the baseline to our best special token setting: language and type tokens for English $\rightarrow$ X, and language, type and script tokens for X $\rightarrow$ English.
Our null hypothesis was that there is no difference between the medians of the two groups.
% Although we are primarily interested in detecting performance improvements, we used a two-tailed alternative hypothesis, i.e. that difference from the baseline is different from 0, to identify cases where performance significantly increases or decreases.
In Table~\ref{best-results}, we use boldface to indicate where significant deviations from the language-only token baseline were observed and where a statistically significant result was obtained at the $p < 0.05$ level.

% TODO: Reconsider putting this back in when revised
% Finally, we note that all the results above are reported on data taken directly from our resource. While this could be seen as a problem, we feel that the preprocessing we apply to the raw Wikidata names is fairly minimal and thus believe there is not a large difference between evaluating on ``gold-standard'' entities and our test set.

\section{Ethics and broader impact}

We believe that the creation of this resource will benefit the speakers of the included languages by enabling improvements to language technology and access to information in more languages.
This resource consists only of information voluntarily provided to a user-edited database regarding notable entities, and does not include data collected from sources like social media that users did not know would become part of a public dataset.

However, like any language technology resource, this work could have unanticipated negative impact, and this impact could be magnified because some of this resource contains data in the languages of marginalized and minoritized populations.

A potential risk in using this resource is that quality issues in Wikidata can be passed to downstream systems, resulting in unexpectedly poor performance.
As an extreme example of this, much of the content of Scots Wikipedia and associated content in Wikidata was found to have been created or edited by someone with minimal proficiency in the language,\footnote{\href{https://www.theguardian.com/uk-news/2020/aug/26/shock-an-aw-us-teenager-wrote-huge-slice-of-scots-wikipedia}{Shock an aw: US teenager wrote huge slice of Scots Wikipedia, \emph{The Guardian}, August 26th 2020.}} and this data was used in the training of Multilingual BERT \citep{devlin-etal-2019-bert}.
We encourage users of this resource who build systems to collaborate with native speakers to verify data quality in the specific languages used.

\section{Conclusion}

ParaNames enables the modeling of names cross-linguistically for millions of entities in over 400 languages.
While we use Wikidata as our source, we have not simply taken its data as-is.
Through careful analysis of the source data, we have developed an approach to processing it to create a massively multilingual name corpus where names are in the expected scripts and all entities have usable entity type information.
We do not claim that this resource will provide perfect data.
However, to the best of our knowledge it does provide the broadest coverage of entities and languages available of any resource to date.
The release of this resource enables multifaceted research in names, including name translation/transliteration and named entity recognition and linking, especially in lower-resourced languages.

In addition to describing our process for creating this resource, we have performed experiments for a canonical name translation task enabled by it.
We have demonstrated the value of providing entity type information in this task and established that while for some languages a current off-the-shelf model can perform relatively well, for many languages there is much room for improvement.
While our experiments have been constrained by the computational resources available to us, we believe an important area for future work is to use more advanced models to perform the canonical name translation task and to do so at larger scale, including more languages in the models.

% In our experiments, we have quantified to what extent transliteration models can benefit from being trained on massively multilingual collections of potentially noisy entity names, as opposed to more curated, smaller datasets used in past work.
% While this process inevitably exposes the model to more noise, we believe that using a larger dataset like ParaNames can help us understand how robust neural transliteration models are to such variations.
% While our model works remarkably well for some languages, the canonical name translation is far from solved as the large between-language variation in accuracy and CER shows.

% We believe that the broader impact of this work will be the improvements to applications that it enables and the scrutiny it can place on the contents of Wikidata so that the quantity and quality of entity names across languages can be improved.

%%%% COMBINED BIG TABLE
\begin{landscape}
\begin{table}
% \footnotesize
\centering
\resizebox{9.75in}{!}{
\begin{tabular}{lrrrrrrrrrrrrrrrrrr}
\toprule
           & \multicolumn{3}{c}{None} & \multicolumn{3}{c}{Script only} & \multicolumn{3}{c}{Language only} & \multicolumn{3}{c}{Language + script} & \multicolumn{3}{c}{Language + type} & \multicolumn{3}{c}{Language + type + script} \\
           \cmidrule(lr){2-4} \cmidrule(lr){5-7} \cmidrule(lr){8-10} \cmidrule(lr){11-13} \cmidrule(lr){14-16} \cmidrule(lr){17-19}
           &        Accuracy &             CER &              F1 &        Accuracy &             CER &              F1 &        Accuracy &             CER &              F1 &        Accuracy &             CER &              F1 &        Accuracy &             CER &              F1 &         Accuracy &             CER &              F1 \\
\midrule
     Arabic & 25.46 $\pm$ .07 &  0.32 $\pm$ .00 & 89.23 $\pm$ .01 & 25.54 $\pm$ .04 &  0.32 $\pm$ .00 & 89.19 $\pm$ .02 & 25.74 $\pm$ .03 &  0.32 $\pm$ .00 & 89.23 $\pm$ .01 & 26.00 $\pm$ .04 &  0.32 $\pm$ .00 & 89.28 $\pm$ .01 & 27.24 $\pm$ .01 &  0.31 $\pm$ .00 & 89.54 $\pm$ .01 &  27.07 $\pm$ .05 &  0.31 $\pm$ .00 & 89.51 $\pm$ .01 \\
     Greek & 24.82 $\pm$ .03 &  0.36 $\pm$ .00 & 86.57 $\pm$ .00 & 24.98 $\pm$ .02 &  0.35 $\pm$ .00 & 86.66 $\pm$ .01 & 24.70 $\pm$ .03 &  0.35 $\pm$ .00 & 86.60 $\pm$ .01 & 24.84 $\pm$ .03 &  0.35 $\pm$ .00 & 86.70 $\pm$ .01 & 25.72 $\pm$ .06 &  0.35 $\pm$ .00 & 86.80 $\pm$ .01 &  25.74 $\pm$ .08 &  0.35 $\pm$ .00 & 86.81 $\pm$ .01 \\
   Persian & 28.78 $\pm$ .06 &  0.27 $\pm$ .00 & 89.94 $\pm$ .01 & 29.22 $\pm$ .03 &  0.27 $\pm$ .00 & 89.98 $\pm$ .01 & 28.68 $\pm$ .05 &  0.28 $\pm$ .00 & 89.84 $\pm$ .02 & 28.75 $\pm$ .06 &  0.27 $\pm$ .00 & 89.90 $\pm$ .01 & 29.49 $\pm$ .05 &  0.27 $\pm$ .00 & 89.90 $\pm$ .02 &  29.47 $\pm$ .04 &  0.27 $\pm$ .00 & 89.92 $\pm$ .02 \\
    Hebrew & 15.02 $\pm$ .03 &  0.44 $\pm$ .00 & 84.52 $\pm$ .01 & 15.37 $\pm$ .02 &  0.43 $\pm$ .00 & 84.62 $\pm$ .01 & 15.24 $\pm$ .07 &  0.44 $\pm$ .00 & 84.58 $\pm$ .02 & 15.16 $\pm$ .03 &  0.44 $\pm$ .00 & 84.58 $\pm$ .01 & 16.38 $\pm$ .02 &  0.42 $\pm$ .00 & 84.93 $\pm$ .01 &  16.34 $\pm$ .03 &  0.42 $\pm$ .00 & 84.89 $\pm$ .01 \\
  Armenian & 39.81 $\pm$ .06 &  0.28 $\pm$ .00 & 90.05 $\pm$ .00 & 39.68 $\pm$ .06 &  0.28 $\pm$ .00 & 90.01 $\pm$ .01 & 39.92 $\pm$ .05 &  0.28 $\pm$ .00 & 90.04 $\pm$ .01 & 39.84 $\pm$ .02 &  0.27 $\pm$ .00 & 90.06 $\pm$ .01 & 40.33 $\pm$ .03 &  0.27 $\pm$ .00 & 90.11 $\pm$ .00 &  40.70 $\pm$ .08 &  0.27 $\pm$ .00 & 90.14 $\pm$ .01 \\
  Japanese & 28.65 $\pm$ .03 &  0.33 $\pm$ .00 & 87.71 $\pm$ .01 & 28.89 $\pm$ .07 &  0.33 $\pm$ .00 & 87.81 $\pm$ .01 & 29.00 $\pm$ .04 &  0.33 $\pm$ .00 & 87.79 $\pm$ .01 & 28.96 $\pm$ .04 &  0.33 $\pm$ .00 & 87.83 $\pm$ .01 & 29.21 $\pm$ .02 &  0.32 $\pm$ .00 & 87.88 $\pm$ .01 &  29.53 $\pm$ .04 &  0.32 $\pm$ .00 & 87.90 $\pm$ .01 \\
  Georgian & 34.52 $\pm$ .03 &  0.29 $\pm$ .00 & 89.26 $\pm$ .01 & 34.66 $\pm$ .03 &  0.29 $\pm$ .00 & 89.31 $\pm$ .01 & 34.44 $\pm$ .02 &  0.29 $\pm$ .00 & 89.29 $\pm$ .01 & 34.61 $\pm$ .06 &  0.29 $\pm$ .00 & 89.33 $\pm$ .01 & 35.43 $\pm$ .02 &  0.29 $\pm$ .00 & 89.34 $\pm$ .01 &  35.56 $\pm$ .05 &  0.29 $\pm$ .00 & 89.40 $\pm$ .01 \\
    Kazakh & 56.06 $\pm$ .03 &  0.16 $\pm$ .00 & 94.04 $\pm$ .00 & 55.91 $\pm$ .04 &  0.16 $\pm$ .00 & 94.02 $\pm$ .00 & 55.38 $\pm$ .04 &  0.16 $\pm$ .00 & 93.93 $\pm$ .01 & 55.41 $\pm$ .06 &  0.16 $\pm$ .00 & 93.93 $\pm$ .01 & 56.44 $\pm$ .03 &  0.16 $\pm$ .00 & 94.02 $\pm$ .01 &  56.33 $\pm$ .05 &  0.16 $\pm$ .00 & 94.01 $\pm$ .00 \\
    Korean & 33.74 $\pm$ .03 &  0.32 $\pm$ .00 & 88.57 $\pm$ .01 & 33.53 $\pm$ .05 &  0.32 $\pm$ .00 & 88.54 $\pm$ .02 & 33.27 $\pm$ .05 &  0.32 $\pm$ .00 & 88.46 $\pm$ .01 & 33.58 $\pm$ .04 &  0.32 $\pm$ .00 & 88.59 $\pm$ .01 & 35.22 $\pm$ .03 &  0.31 $\pm$ .00 & 88.97 $\pm$ .01 &  35.20 $\pm$ .04 &  0.31 $\pm$ .00 & 88.98 $\pm$ .01 \\
Lithuanian & 46.21 $\pm$ .02 &  0.29 $\pm$ .00 & 89.34 $\pm$ .01 & 46.46 $\pm$ .02 &  0.28 $\pm$ .00 & 89.39 $\pm$ .01 & 47.39 $\pm$ .03 &  0.28 $\pm$ .00 & 89.53 $\pm$ .01 & 47.51 $\pm$ .03 &  0.28 $\pm$ .00 & 89.58 $\pm$ .01 & 47.81 $\pm$ .05 &  0.28 $\pm$ .00 & 89.66 $\pm$ .01 &  48.06 $\pm$ .06 &  0.27 $\pm$ .00 & 89.72 $\pm$ .01 \\
   Latvian & 66.01 $\pm$ .03 &  0.15 $\pm$ .00 & 94.88 $\pm$ .01 & 66.08 $\pm$ .04 &  0.15 $\pm$ .00 & 94.89 $\pm$ .01 & 67.86 $\pm$ .02 &  0.14 $\pm$ .00 & 95.19 $\pm$ .01 & 68.16 $\pm$ .05 &  0.14 $\pm$ .00 & 95.22 $\pm$ .01 & 68.69 $\pm$ .05 &  0.14 $\pm$ .00 & 95.27 $\pm$ .01 &  68.64 $\pm$ .03 &  0.14 $\pm$ .00 & 95.28 $\pm$ .01 \\
   Russian & 32.95 $\pm$ .04 &  0.38 $\pm$ .00 & 84.78 $\pm$ .01 & 32.86 $\pm$ .03 &  0.38 $\pm$ .00 & 84.75 $\pm$ .01 & 32.81 $\pm$ .06 &  0.38 $\pm$ .00 & 84.80 $\pm$ .02 & 33.16 $\pm$ .06 &  0.38 $\pm$ .00 & 84.87 $\pm$ .02 & 33.17 $\pm$ .04 &  0.38 $\pm$ .00 & 84.82 $\pm$ .01 &  32.92 $\pm$ .04 &  0.38 $\pm$ .00 & 84.82 $\pm$ .03 \\
   Swedish & 88.20 $\pm$ .02 &  0.08 $\pm$ .00 & 97.13 $\pm$ .01 & 88.12 $\pm$ .00 &  0.08 $\pm$ .00 & 97.09 $\pm$ .00 & 88.25 $\pm$ .02 &  0.08 $\pm$ .00 & 97.15 $\pm$ .01 & 88.20 $\pm$ .03 &  0.08 $\pm$ .00 & 97.13 $\pm$ .01 & 88.18 $\pm$ .02 &  0.08 $\pm$ .00 & 97.10 $\pm$ .01 &  88.17 $\pm$ .03 &  0.08 $\pm$ .00 & 97.13 $\pm$ .01 \\
     Tajik & 47.34 $\pm$ .05 &  0.21 $\pm$ .00 & 92.51 $\pm$ .01 & 47.34 $\pm$ .06 &  0.21 $\pm$ .00 & 92.51 $\pm$ .00 & 49.62 $\pm$ .05 &  0.20 $\pm$ .00 & 92.77 $\pm$ .01 & 49.91 $\pm$ .05 &  0.20 $\pm$ .00 & 92.79 $\pm$ .01 & 50.03 $\pm$ .03 &  0.20 $\pm$ .00 & 92.81 $\pm$ .01 &  50.36 $\pm$ .03 &  0.20 $\pm$ .00 & 92.84 $\pm$ .01 \\
      Thai & 44.12 $\pm$ .04 &  0.29 $\pm$ .00 & 89.94 $\pm$ .02 & 43.89 $\pm$ .05 &  0.29 $\pm$ .00 & 89.89 $\pm$ .01 & 43.94 $\pm$ .05 &  0.29 $\pm$ .00 & 89.91 $\pm$ .01 & 44.03 $\pm$ .04 &  0.29 $\pm$ .00 & 89.95 $\pm$ .02 & 45.35 $\pm$ .07 &  0.28 $\pm$ .00 & 90.10 $\pm$ .01 &  45.38 $\pm$ .07 &  0.28 $\pm$ .00 & 90.12 $\pm$ .02 \\
      Urdu & 30.79 $\pm$ .04 &  0.24 $\pm$ .00 & 91.06 $\pm$ .01 & 30.65 $\pm$ .04 &  0.23 $\pm$ .00 & 91.11 $\pm$ .01 & 31.92 $\pm$ .03 &  0.23 $\pm$ .00 & 91.48 $\pm$ .01 & 32.06 $\pm$ .07 &  0.22 $\pm$ .00 & 91.51 $\pm$ .01 & 32.31 $\pm$ .06 &  0.22 $\pm$ .00 & 91.59 $\pm$ .01 &  32.74 $\pm$ .04 &  0.22 $\pm$ .00 & 91.62 $\pm$ .01 \\
Vietnamese & 80.83 $\pm$ .03 &  0.17 $\pm$ .00 & 94.07 $\pm$ .01 & 80.87 $\pm$ .02 &  0.17 $\pm$ .00 & 94.07 $\pm$ .01 & 80.75 $\pm$ .02 &  0.17 $\pm$ .00 & 94.08 $\pm$ .01 & 80.85 $\pm$ .03 &  0.17 $\pm$ .00 & 94.07 $\pm$ .02 & 80.71 $\pm$ .03 &  0.17 $\pm$ .00 & 94.05 $\pm$ .01 &  80.96 $\pm$ .01 &  0.17 $\pm$ .00 & 94.11 $\pm$ .01 \\
\midrule
   Overall & 42.55 $\pm$ .01 &  0.27 $\pm$ .00 & 90.21 $\pm$ .00 & 42.59 $\pm$ .01 &  0.27 $\pm$ .00 & 90.23 $\pm$ .00 & 42.88 $\pm$ .02 &  0.27 $\pm$ .00 & 90.27 $\pm$ .01 & 43.00 $\pm$ .01 &  0.27 $\pm$ .00 & 90.31 $\pm$ .00 & 43.63 $\pm$ .01 &  0.27 $\pm$ .00 & 90.40 $\pm$ .00 &  43.72 $\pm$ .02 &  0.27 $\pm$ .00 & 90.42 $\pm$ .00 \\
\midrule
    Arabic &               - &               - &               - &               - &               - &               - & 41.70 $\pm$ .02 &  0.28 $\pm$ .00 & 89.40 $\pm$ .01 &               - &               - &               - & 41.67 $\pm$ .03 &  0.28 $\pm$ .00 & 89.33 $\pm$ .01 &                - &               - &               - \\
     Greek &               - &               - &               - &               - &               - &               - & 29.67 $\pm$ .06 &  0.36 $\pm$ .00 & 86.88 $\pm$ .01 &               - &               - &               - & 30.18 $\pm$ .03 &  0.36 $\pm$ .00 & 86.88 $\pm$ .01 &                - &               - &               - \\
   Persian &               - &               - &               - &               - &               - &               - & 22.90 $\pm$ .05 &  0.41 $\pm$ .00 & 81.64 $\pm$ .05 &               - &               - &               - & 22.60 $\pm$ .07 &  0.42 $\pm$ .00 & 81.81 $\pm$ .05 &                - &               - &               - \\
    Hebrew &               - &               - &               - &               - &               - &               - & 35.71 $\pm$ .03 &  0.34 $\pm$ .00 & 88.16 $\pm$ .01 &               - &               - &               - & 36.00 $\pm$ .03 &  0.33 $\pm$ .00 & 88.23 $\pm$ .01 &                - &               - &               - \\
  Armenian &               - &               - &               - &               - &               - &               - & 50.45 $\pm$ .05 &  0.22 $\pm$ .00 & 92.41 $\pm$ .01 &               - &               - &               - & 51.78 $\pm$ .06 &  0.21 $\pm$ .00 & 92.43 $\pm$ .01 &                - &               - &               - \\
  Japanese &               - &               - &               - &               - &               - &               - & 28.70 $\pm$ .01 &  0.42 $\pm$ .00 & 84.42 $\pm$ .02 &               - &               - &               - & 28.61 $\pm$ .04 &  0.42 $\pm$ .00 & 84.64 $\pm$ .01 &                - &               - &               - \\
  Georgian &               - &               - &               - &               - &               - &               - & 51.82 $\pm$ .04 &  0.22 $\pm$ .00 & 92.56 $\pm$ .01 &               - &               - &               - & 53.13 $\pm$ .05 &  0.22 $\pm$ .00 & 92.68 $\pm$ .01 &                - &               - &               - \\
    Kazakh &               - &               - &               - &               - &               - &               - & 58.69 $\pm$ .09 &  0.14 $\pm$ .00 & 94.85 $\pm$ .02 &               - &               - &               - & 59.78 $\pm$ .07 &  0.13 $\pm$ .00 & 95.00 $\pm$ .01 &                - &               - &               - \\
    Korean &               - &               - &               - &               - &               - &               - & 38.63 $\pm$ .05 &  0.33 $\pm$ .00 & 88.18 $\pm$ .01 &               - &               - &               - & 39.29 $\pm$ .04 &  0.33 $\pm$ .00 & 88.34 $\pm$ .01 &                - &               - &               - \\
Lithuanian &               - &               - &               - &               - &               - &               - & 50.76 $\pm$ .09 &  0.23 $\pm$ .00 & 91.61 $\pm$ .03 &               - &               - &               - & 54.21 $\pm$ .05 &  0.20 $\pm$ .00 & 92.62 $\pm$ .02 &                - &               - &               - \\
   Latvian &               - &               - &               - &               - &               - &               - & 69.28 $\pm$ .07 &  0.13 $\pm$ .00 & 95.49 $\pm$ .01 &               - &               - &               - & 70.50 $\pm$ .04 &  0.12 $\pm$ .00 & 95.79 $\pm$ .00 &                - &               - &               - \\
   Russian &               - &               - &               - &               - &               - &               - & 44.59 $\pm$ .04 &  0.33 $\pm$ .00 & 89.81 $\pm$ .02 &               - &               - &               - & 45.68 $\pm$ .03 &  0.32 $\pm$ .00 & 89.94 $\pm$ .01 &                - &               - &               - \\
   Swedish &               - &               - &               - &               - &               - &               - & 85.60 $\pm$ .04 &  0.10 $\pm$ .00 & 96.11 $\pm$ .02 &               - &               - &               - & 85.69 $\pm$ .02 &  0.10 $\pm$ .00 & 96.13 $\pm$ .01 &                - &               - &               - \\
     Tajik &               - &               - &               - &               - &               - &               - & 54.38 $\pm$ .02 &  0.18 $\pm$ .00 & 93.82 $\pm$ .02 &               - &               - &               - & 54.83 $\pm$ .04 &  0.17 $\pm$ .00 & 93.99 $\pm$ .01 &                - &               - &               - \\
      Thai &               - &               - &               - &               - &               - &               - & 14.80 $\pm$ .04 &  0.42 $\pm$ .00 & 83.01 $\pm$ .02 &               - &               - &               - & 15.07 $\pm$ .05 &  0.41 $\pm$ .00 & 83.43 $\pm$ .02 &                - &               - &               - \\
      Urdu &               - &               - &               - &               - &               - &               - & 14.14 $\pm$ .08 &  0.45 $\pm$ .00 & 80.74 $\pm$ .03 &               - &               - &               - & 14.23 $\pm$ .06 &  0.45 $\pm$ .00 & 80.76 $\pm$ .03 &                - &               - &               - \\
Vietnamese &               - &               - &               - &               - &               - &               - & 48.86 $\pm$ .01 &  0.35 $\pm$ .00 & 82.87 $\pm$ .01 &               - &               - &               - & 48.78 $\pm$ .03 &  0.35 $\pm$ .00 & 82.89 $\pm$ .01 &                - &               - &               - \\
\midrule
   Overall &               - &               - &               - &               - &               - &               - & 43.57 $\pm$ .02 &  0.29 $\pm$ .00 & 88.94 $\pm$ .01 &               - &               - &               - & 44.24 $\pm$ .01 &  0.29 $\pm$ .00 & 89.11 $\pm$ .00 &                - &               - &               - \\
\bottomrule
\end{tabular}
}
\caption{Canonical name translation results for the X $\rightarrow$ English (top) and English $\rightarrow$ X directions across different special token settings.}
\label{bigtable-all-results}
\end{table}
\end{landscape}

% Entries for the entire Anthology, followed by custom entries
\bibliography{anthology,custom}
\bibliographystyle{acl_natbib}

\end{document}